\documentclass[sigconf]{acmart}

\AtBeginDocument{%
  \providecommand\BibTeX{{%
    \normalfont B\kern-0.5em{\scshape i\kern-0.25em b}\kern-0.8em\TeX}}}

\setcopyright{acmcopyright}
\copyrightyear{2018}
\acmYear{2018}
\acmDOI{XXXXXXX.XXXXXXX}

\acmConference[Conference acronym 'XX]{Make sure to enter the correct
  conference title from your rights confirmation emai}{June 03--05,
  2018}{Woodstock, NY}
\acmPrice{15.00}
\acmISBN{978-1-4503-XXXX-X/18/06}

\usepackage{threeparttable}
\usepackage{multirow}

\begin{document}

\title{Empowering Many, Biasing a Few: Generalist Credit Scoring through Large Language Models}

\author{Duanyu Feng}
\authornote{Both authors contributed equally to this research.}
\email{fengduanyu@stu.scu.edu.cn}
\author{Yongfu Dai}
\authornotemark[1]
\email{wal.daishen@gmail.com}
\affiliation{%
  \institution{Sichuan University}
  \country{China}
}

\author{Jimin Huang}
\email{ jimin@chancefocus.com}
\affiliation{%
  \institution{ChanceFocus (Shanghai) AMC. }
  \country{China}
  }

\author{Yifang Zhang}
 \email{ zhangyf_ivy@foxmail.com}
\affiliation{%
 \institution{Sichuan University}
  \country{China}
  }

\author{Qianqian Xie}
  \email{xqq.sincere@gmail.com}
\affiliation{%
  \institution{Wuhan University}
  \country{China}
}

\author{Weiguang Han}
\email{ han.wei.guang@whu.edu.cn}
\affiliation{%
  \institution{Wuhan University}
  \country{China}
}

\author{Zhengyu Chen}
\email{2019302120293@whu.edu.cn}
\affiliation{%
  \institution{Wuhan University}
  \country{China}
}

\author{Alejandro Lopez-Lira}
\email{ alejandro.lopez-lira@warrington.ufl.edu}
\affiliation{%
  \institution{University of Florida}
  \country{USA}
}

\author{Hao Wang}
\authornote{Corresponding Author.}
\email{wangh@scu.edu.cn}
\affiliation{%
 \institution{Sichuan University}
  \country{China}
  }

\renewcommand{\shortauthors}{Feng and Dai, et al.}

\begin{abstract}
In the financial industry, credit scoring is a fundamental element, shaping access to credit and determining the terms of loans for individuals and businesses alike. Traditional credit scoring methods, however, often grapple with challenges such as narrow knowledge scope and isolated evaluation of credit tasks. Our work posits that Large Language Models (LLMs) have great potential for credit scoring tasks, with strong generalization ability across multiple tasks.
To systematically explore LLMs for credit scoring, we propose the first open-source comprehensive framework.
We curate a novel benchmark covering 9 datasets with 14K samples, tailored for credit assessment and a critical examination of potential biases within LLMs, and the novel instruction tuning data with over 45k samples. We then propose the first Credit and Risk Assessment Large Language Model (CALM) by instruction tuning, tailored to the nuanced demands of various financial risk assessment tasks. We evaluate CALM, existing state-of-art (SOTA) methods, open source and closed source LLMs on the build benchmark. Our empirical results illuminate the capability of LLMs to not only match but surpass conventional models, pointing towards a future where credit scoring can be more inclusive, comprehensive, and unbiased. We contribute to the industry's transformation by sharing our pioneering instruction-tuning datasets, credit and risk assessment LLM, and benchmarks with the research community and the financial industry \footnote{\url{https://github.com/colfeng/CALM}}.
\end{abstract}

%
%

\ccsdesc[500]{Social and professional topics~Economic impact}
\ccsdesc[500]{Social and professional topics~Privacy policies}

\ccsdesc[500]{Computing methodologies~Language resources}
\ccsdesc[300]{Computing methodologies~Natural language generation}

\keywords{Credit Scoring, Large Language Models, Bias Analysis}



\maketitle

\section{Introduction}
Credit and risk assessment is vital in the financial industry, determining the probability of repayment by borrowers, from individuals to nations \cite{cao2021data,jeong2022customs}. These evaluations, critical for maintaining financial stability, have increasingly moved online.
Companies now use online methods for individual assessments like credit scoring and claim analysis to predict default risks \cite{dastile2020statistical} and ensure equitable claim resolutions \cite{rawat2021application}, relying on historical account data and online application details. For community-level protection, institutions deploy online fraud detection \cite{bhatore2020machine} to safeguard against illicit financial activities. Similarly, tools for detecting financial distress help in preempting economic downturns, influencing wider investment and policy decisions \cite{cao2020ai}. These varied online assessments are pivotal, shaping not only personal financial outcomes but also global economic health.

Existing methods in financial credit and risk assessment, are often rule-based or machine learning based expert systems, which show limited flexibility across tasks \cite{soui2019rule, runchi2023ensemble, kamaruddin2021egrnn++, wang2022deep, lai2023default}.
These methods are designed specifically for a singular task, 
struggle to generalize or integrate knowledge from different tasks.
For example, when training a Credit Scoring model, it can only be used for that specific task and only utilize relevant features related to a particular task \cite{balasubramanian2022substituting}, making them unsuitable for other tasks.
Furthermore, these approaches miss out on the advantages of transferring insights between financial activities \cite{vandenhende2020revisiting, shen2022reject}.
Skills used in claim analysis, such as anti-fraud techniques from insurance in banking, can also be applied to various financial activities such as lending and borrowing. 
There is a clear demand for a generalist approach in credit scoring that can navigate various financial tasks effectively, drawing from a broad knowledge base to enhance predictive accuracy and model adaptability \cite{kou2019machine}.

Recently, the advent of Large Language Models (LLMs) presents an opportunity to transcend these limitations through multi-task learning and few-shot generalization \cite{sanh2021multitask}.
The integration of LLMs into financial assessments is gaining traction, with research exploring how these models can identify task correlations and generalize across financial tasks \cite{xie2023pixiu, xie2023wall, wu2023bloomberggpt, zhang2023unveiling}, potentially marking a paradigm shift in credit and risk evaluation methodologies.

Despite their potential, the application of LLMs to credit and risk assessment is not without challenges. The process often involves analyzing tabular data, which contains symbolic information that is markedly different from the natural language typically processed by LLMs, and current LLMs' performance with such data remains limited \cite{li2023tablegpt}. Moreover, issues such as class imbalance in financial datasets and the need to avoid bias in sensitive attributes like age or gender present significant hurdles for training LLMs effectively \cite{hanafy2021machine, casper2023open,wolf2023fundamental, ferrer2021bias}. 
It is unclear how LLMs may navigate the specific challenges of the financial industry's credit and risk assessments. To clarify this, we delve into several key hypotheses:
\begin{itemize}
    \item \textbf{H1:} Based on LLMs' broad pretraining, can LLMs overcome the limitations of traditional expert systems by applying their extensive pretraining to diverse online credit and risk assessment tasks, effectively utilizing a wider range of financial knowledge in the process?
    \item \textbf{H2:} Can LLMs, fine-tuned with credit and risk assessment datasets, generalize their learning to understand and manage multiple related credit tasks, potentially developing the generalization ability?
    \item \textbf{H3:} Do the advancements in model capabilities of LLMs come at the cost of fairness and equality in financial decision-making?
\end{itemize}

To assess the potential of LLMs in credit and risk assessment, we established a comprehensive benchmark aligned with our research hypotheses. Addressing \textbf{H1}, we compiled a diverse collection of datasets, amassing over 14K samples that challenge the LLMs to process various types of credit and risk-related tasks. The results indicate that models like GPT-4 \cite{openai2023gpt4} can indeed mitigate the issue of narrow expertise found in traditional systems, adapting to different tasks through targeted prompting.
For \textbf{H2}, we propose the Credit and Risk Assessment Language Model (CALM), employing over 45k instruction tuning data for fine-tuning. CALM's performance suggests that LLMs can transcend the limitations of traditional task-specific models, facilitating knowledge transfer across a spectrum of financial tasks and potentially revolutionizing credit assessment processes.
However, as we delved into \textbf{H3}, we discovered that despite their analytical prowess, LLMs are not immune to biases, highlighting the necessity for vigilant ethical oversight. Ensuring that LLMs are deployed responsibly in financial decision-making processes is paramount to prevent the perpetuation of existing societal biases.

We highlight our contributions as follows:
\begin{itemize} 
    \item We have established the first comprehensive framework for credit scoring using LLMs, which encompasses a curated instruction tuning dataset, specialized LLMs, and a set of benchmarks. This framework is not only a research milestone but also a practical toolset that can be directly applied within the financial industry to enhance the accuracy and depth of credit analysis.
    \item Our research sheds light on the potential of LLMs to enhance the field of credit and risk assessment. By demonstrating their ability to understand and analyze complex financial data, we provide evidence that LLMs could significantly improve the accuracy and efficiency of credit scoring practices in the financial industry.
    \item We bring attention to the ethical considerations inherent in deploying LLMs, particularly in sensitive applications like credit scoring that have far-reaching societal impacts. To promote ethical use and continuous innovation, we have open-sourced all our resources, encouraging scrutiny, adaptation, and advancement of our work within the research community and industry at large.
\end{itemize}

\begin{table*}[!htb]
\centering
\caption{The statistics of the datasets. The "Test/Train/Raw" shows the division of our benchmark, instruction tuning data, and raw data quantities. The number of features in these datasets is presented as "Columns". The "Anonymized" indicates whether the dataset has been transformed into meaningless symbols.}
\vspace{-3mm}
\label{tab:sta_data}
\resizebox{0.99\textwidth}{!}{%
    \begin{tabular}{llcccc}
    \toprule
    Task  & Dataset & Test/Train/Raw & Columns & Anonymized & License \\
    \midrule
    \multirow{3}[2]{*}{Credit Scoring} & German & 300/700/1,000  & 20    & No    & CC BY 4.0 \\
          & Australia & 207/483/690 & 14 & Yes &  CC BY 4.0  \\
          & Lending Club &  4,036/ - /1,345,310 & 21 &  No  &  CC0 1.0 \\
    \midrule
    \multirow{2}[2]{*}{Fraud Detection} & Credit Card Fraud & 3,418/7,974/284,807 & 29 &  Yes & (DbCL) v1.0  \\
          & ccFraud & 3,145/7,340/1,048,575 & 7  &  No   &  Public  \\
    \midrule
    \multirow{2}[2]{*}{Financial Distress Identification} & Polish & 2,604/ - /43,405 & 64  &  No   & CC BY 4.0 \\
          & Taiwan Economic Journal &   2,046/4,773/6,819    &   95   &  No   & CC BY 4.0 \\
    \midrule
    \multirow{2}[2]{*}{Claim Analysis} & PortoSeguro &  3,571/ - /595,212   &  57   &  Yes  &  Public \\
          & Travel Insurance & 3,800/8,865/63,326  & 9   &   No  & (ODbL) v1.0 \\
    \bottomrule
    \end{tabular}%
}
\vspace{-4.5mm}
\end{table*}

\vspace{-2mm}
\section{Related Works}
\textbf{Credit and risk assessment.}
Currently, credit and risk assessment has a significant impact on finance and society, and online credit assessment services have taken over a major part of the tasks in this field.
To solve these tasks, most of the companies still use expert systems \cite{abror2023bankruptcy, jemai2023feature}. They collect the data and introduce a priori knowledge in feature engineering as well as in the modeling phase. 
However, when it comes to modeling, various companies may have different focuses. Some companies care more about the final results. They design complex neural networks \cite{benchaji2021enhanced, wang2022deep} or ensemble methods like XGBoost \cite{schmitt2022deep, jammalamadaka2023responsible, mokheleli2023machine}, Random-Forest and iForest \cite{hanafy2021machine, li2023exploring} to more effectively find the nonlinear connection between features and labels and improve models' performance.
Some data sampling like SMOTE-based methods \cite{aly2022intelligent, abror2023bankruptcy, muslim2023ensemble} are also used in most of the works to solve the imbalanced problem. 
Other companies may more care interpretability and transparency of the methods to meet customer and regulatory needs \cite{bucker2022transparency}. Therefore, they use the methods like classical rule-dependent models \cite{soui2019rule,maldonado2020credit}, logistic regression\cite{runchi2023ensemble}, and entropy-based approaches \cite{carta2020combined}. However, most of these expert systems lack generalization and correlation as they are designed for specific tasks and datasets, which exacerbates a narrow knowledge scope and isolated tasks, requiring significant redesign and redevelopment \cite{balasubramanian2022substituting}.

Despite the performance and transparency, researchers also investigate the effectiveness of expert systems in handling sensitive data such as \textit{age}, \textit{gender}, and \textit{ethnicity} \cite{ahelegbey2019latent,bussmann2021explainable}. They make efforts to prevent expert systems methods from producing discriminatory results or to eliminate biased effects after making predictions \cite{jammalamadaka2023responsible}. As such, we consider using LLMs that can store the general knowledge and learning domain knowledge \cite{meng2022locating, zhen2022survey} to explore these online tasks in credit and risk assessment.
We also take into account the potential bias issues when utilizing LLMs to investigate task abilities, as LLMs are susceptible to some biases \cite{venkit2023nationality, salewski2023context}.

\textbf{LLMs for financial and the evaluation benchmark.} 
Although LLMs have been applied in many areas of finance \cite{xie2023pixiu, zhang2023unveiling, lopez2023can}, to our best knowledge, there is limited research using LLMs to explore credit and risk assessment.
Before the release of ChatGPT \footnote{\url{https://openai.com/blog/chatgpt}}, a highly anticipated LLM, studies were mainly conducted using BERT-based pre-trained language models (PLMs) \cite{devlin2018bert} which are smaller in size compared to current LLMs. 
The financial pre-trained language models like finBERT \cite{araci2019finbert}, FinBERT \cite{yang2020finbert}, and FLANG \cite{shah2022flue} are proposed as the backbone for the NLP-related tasks in the financial domain \footnote{There are some similar financial PLMs also called finBERT/FinBERT \cite{desola2019finbert,liu2021finbert,huang2023finbert}.}. 
They demonstrated that PLMs have a strong ability to solve some financial tasks, such as financial sentiment analysis, financial named entity recognition, and question answering \cite{shah2022flue}.
Recently, with the emergence of LLMs such as ChatGPT and GPT-4 \cite{openai2023gpt4}, which have better text processing capabilities, there is growing interest in their performance in financial tasks \cite{xie2023pixiu, ko2023can}.
Many studies have attempted to utilize the powerful general knowledge capabilities of LLMs \cite{meng2022locating,sanh2021multitask} to directly address various tasks in the financial domain \cite{zhang2023unveiling, lopez2023can, ko2023can}.
In addition, some preliminary work has tried to construct financial LLMs that can tackle multi-tasks specific to the domain, such as BBT-Fin \cite{lu2023bbt}, Bloomberg \cite{wu2023bloomberggpt}, PIXIU \cite{xie2023pixiu} and FinGPT\cite{yang2023fingpt}. 
However, it is worth noting that these studies still pay little attention to problems with tabular data \cite{tavakoli2023multi}, such as credit and risk assessment. 

In terms of evaluating financial LLMs, the current benchmarks are led by PIXIU \cite{xie2023pixiu} and BBT-CFLEB \cite{lu2023bbt}. They have tested most of the LLMs, including Vicuna \cite{chiang2023vicuna}, Llama1,2\cite{touvron2023llama} and GPT-4 \cite{openai2023gpt4}.
Our approach differs from theirs in that we do not aim to construct a benchmark for the entire financial domain. Instead, we take a close look at evaluating the performance of LLMs in credit and risk assessment with its online tasks, and we also take into consideration the potential bias of LLMs in this field. 

\section{Data for Benchmark and instruction tuning}
We develop the first dataset for benchmarking and instruction tuning of large language models (LLMs) in the field of credit and risk assessment. We sourced 9 open-source datasets, rich in complexity to challenge LLMs with diverse, tabular data scenarios, and to identify potential biases. The benchmark comprises all these datasets (14K entries) for a thorough model evaluation. For instruction tuning, we use 6 datasets (initially 30K, expanded to 45K after processing), which exclude 3 datasets to test the LLM's ability to generalize to new tasks. 
This method ensures that the LLM is rigorously tested for both its learning breadth and applicability.

\subsection{Raw Data Collection} 
Following existing work on online credit and task assessment~\cite{bhatore2020machine,cao2020ai}, our study adopts a comprehensive approach, gathering nine open-source datasets that collectively cover the spectrum of four essential task types in credit and risk assessment. Specifically, these task types include \textbf{credit scoring}, \textbf{fraud detection}, \textbf{financial distress identification}, and \textbf{claim analysis}. Each represents a binary classification problem and is underpinned by tabular data, providing a broad, illustrative snapshot of the challenges and complexities inherent in financial decision-making models.
The statistical details of these datasets and the data size for our benchmark and instruction-tuning are shown in Table \ref{tab:sta_data}. 


\subsubsection{\textbf{Credit Scoring.}}
Credit scoring is a critical process used by financial institutions to determine the likelihood that a borrower will repay their debts~\cite{bhatore2020machine}. It involves analyzing the financial information that individuals submit with their credit applications. This analysis helps lenders decide who should receive a loan, what interest rate to charge, and the loan terms. To explore this area, we've chosen three widely recognized datasets, known as \textit{German} \cite{misc_statlog_(german_credit_data)_144}, \textit{Australia} \cite{misc_statlog_(australian_credit_approval)_143}, and \textit{Lending Club}\footnote{\url{https://www.kaggle.com/datasets/wordsforthewise/lending-club}}, which provide detailed insights into the factors that affect an individual's credit score and the risk they pose to lenders.

\textit{German \cite{misc_statlog_(german_credit_data)_144}} is a classic dataset used for credit scoring, consisting of information on 1,000 loan applicants. It provides a mix of 20 attributes per applicant, including 13 qualitative descriptions and 7 quantitative figures. Among these attributes, it captures sensitive personal details such as marital status and gender (\textit{personal status and sex}), age (\textit{age}), and whether the individual is a foreign worker (\textit{foreign worker or not}). For our purposes, we utilize the version of the dataset that includes natural language descriptions, allowing us to explore the nuances of credit scoring in a real-world context.

\textit{Australia \cite{misc_statlog_(australian_credit_approval)_143}} 
 is a credit scoring dataset that stands out because it has been anonymized using symbols instead of actual names for its features. It includes data on 690 individuals, broken down into 8 categorical and 6 numerical attributes, to assess their creditworthiness without revealing their personal information.

\textit{Lending Club}\footnote{\url{https://www.kaggle.com/datasets/wordsforthewise/lending-club}} 
provides a comprehensive look at loan transactions from 2007 to 2018 on the United States' largest peer-to-peer lending platform. It contains around 1.3 million records of borrower data. Following the methods of a previous study \cite{10.1007/978-981-99-2385-4_44}, we focus on 21 specific features of the loan applicants, such as the amount of the installment, the purpose of the loan, and the state of the borrower's address. Our primary variable of interest, \textit{loan status}, is used to categorize the loans into 'Fully Paid' or 'Charged Off', indicating whether the loan was repaid or defaulted, respectively.

\subsubsection{\textbf{Fraud Detection.}}
Fraud detection is a task closely related to credit scoring, where the goal is to identify whether online loan applications are genuine or deceptive. This process is crucial for maintaining the integrity of financial systems and protecting both the institutions and their customers from financial losses. In our research, we have gathered two datasets specifically designed for this task: \textit{Credit Card Fraud}\footnote{\url{https://www.kaggle.com/datasets/mlg-ulb/creditcardfraud}} and \textit{ccFraud \cite{10.1145/2980258.2980319}}. These datasets are notably imbalanced, a common challenge in fraud detection, with the actual cases of fraud representing a very small proportion of the total applications, as 0.17\% and 5.98\% respectively.

\textit{Credit Card Fraud}\footnote{\url{https://www.kaggle.com/datasets/mlg-ulb/creditcardfraud}} 
is a well-known, anonymized collection of data often used for detecting fraudulent activities in financial transactions. It includes 30 features, with 28 of them transformed into non-identifiable symbols through Principal Component Analysis (PCA) to protect sensitive information. Originally, this dataset encompasses a total of 284,807 transaction records.

\textit{ccFraud \cite{10.1145/2980258.2980319}} 
includes around 1 million transaction samples, and for our purposes, we have selected a subset of about 10,000 samples. This dataset contains sensitive information, including the gender of individuals (\textit{gender}), which is a significant feature in our analysis to assess bias.

\subsubsection{\textbf{Financial Distress Identification.}}
This task aims to predict if a company is at risk of or is currently experiencing bankruptcy based on its publicly available online data. This is an important process for stakeholders to understand the financial health and stability of a company~\cite{kuiziniene2022systematic}. For this task, we use two widely recognized datasets, \textit{Polish \cite{misc_polish_companies_bankruptcy_data_365}} and \textit{Taiwan Economic Journal}\footnote{\url{https://www.kaggle.com/datasets/fedesoriano/company-bankruptcy-prediction}}. 

\textit{Polish \cite{misc_polish_companies_bankruptcy_data_365}} is a bankruptcy prediction dataset with 43,405 records for Polish companies. It assesses the bankruptcy status of companies that were still in operation during the years 2007 to 2013. The dataset comprises 64 features, with missing values replaced by 1. Reflecting real-world business conditions, the dataset is significantly imbalanced, with only 3\% of the companies having gone bankrupt. 

\textit{Taiwan Economic Journal}\footnote{\url{https://www.kaggle.com/datasets/fedesoriano/company-bankruptcy-prediction}} 
is a prominent resource in corporate bankruptcy prediction research, encompassing data from 1999 to 2009. It includes a detailed set of 95 features that provide insights into the financial status, performance metrics, and other pertinent details of companies. This dataset contains records for 6,819 companies, with only 4.8\% of these representing instances of bankruptcy.

\subsubsection{\textbf{Claim Analysis.}}
Claim analysis is a critical task for insurance companies, where they analyze claims to identify any fraudulent activity. Fraudulent claims are those that are not genuine and could be attempts to receive payment under false pretenses. Legitimate claims, on the other hand, are valid and honest requests for payment due to losses covered by an insurance policy. This distinction is important to prevent financial losses due to fraud and to ensure that only rightful claims are paid~\cite{RAWAT2021100012}. To study this, we have selected two datasets, \textit{PortoSeguro \cite{hanafy2021machine}} and \textit{Travel Insurance}\footnote{\url{https://www.kaggle.com/datasets/mhdzahier/travel-insurance}}, which provide perspectives from both the beneficiaries that are filing the claims and the insurance companies that are assessing these claims for authenticity. Both datasets are imbalanced, meaning that fraudulent claims are far fewer than legitimate ones, which is a common scenario in real-world insurance claim analysis.

\textit{PortoSeguro \cite{hanafy2021machine}} includes approximately 10,000 records from a Brazilian insurance company, representing a 2\% sample from the original dataset. In this dataset, 57 client features have been anonymized into symbols that don’t directly reveal any identifiable information, ensuring privacy.

\textit{Travel Insurance}\footnote{\url{https://www.kaggle.com/datasets/mhdzahier/travel-insurance}} 
originates from a travel insurance service provider in Singapore. It originally includes 10 features, categorized into 6 categorical and 4 numerical types. However, due to 71\% of the data for the \textit{Gender} feature being missing, we have opted to exclude this feature. We've also condensed the dataset to 20\% of its original volume. The preprocessing aligns with the methodology outlined by \citet{RAWAT2021100012}. It’s important to note that this dataset includes sensitive information such as the age of individuals (\textit{Age}).

\begin{figure*}[!htb]
\vspace{-1mm}
\centering
\includegraphics[width=1.6\columnwidth]{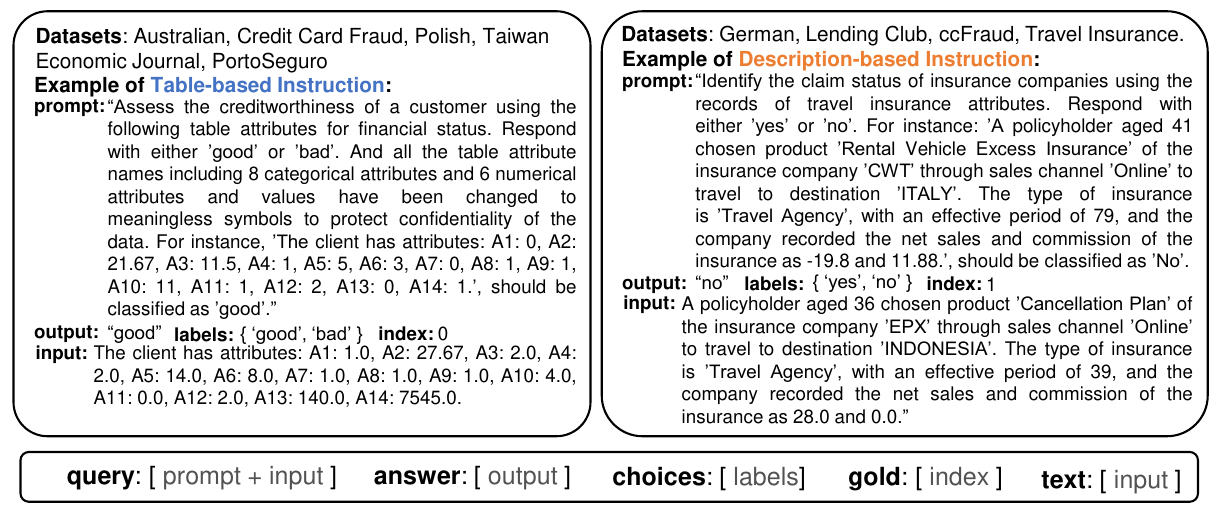} 
\vspace{-4mm}
\caption{The template and example of our instruction data.}
\label{fig3}
 \vspace{-5mm}
\end{figure*}

\subsection{Data Construction}

We then use the above datasets to construct our benchmark and instruction-tuning data. We first construct our prompts similar to the existing instruction data of other financial tasks \cite{xie2023pixiu}. These prompts are then reviewed by financial experts to ensure that the meaning is correct. We also use ChatGPT to confirm that our prompt templates can be answered and also use it to optimize the prompts. 

We develop two types of prompts, \textbf{table-based} and \textbf{description-based}, to evaluate or further fine-tune LLMs, shown in Figure \ref{fig3}. This is because some datasets contain numerous features and meaningless symbols.
Both forms of prompts follow a similar template. For instance, [prompt] is a prompt created for each data, [input] expresses a single data from the original datasets, [output] is the corresponding correct label for the [input], [choices] are the available label options (such as [``Yes", ``No"] or [``good", ``bad"]), and [index] is the index of the correct label in the label set. 
The [choices] and [index] are the special parts of our benchmark, which are further used as parts of the template to evaluate the LLMs and do not exist in our instruction-tuning data.
Therefore, the differences between the two prompts come from the specific construction of [prompt] and [input]. 
Strictly speaking, we provide a very simple example that may have omissions in [prompt], so the datasets are one-shot or nearly zero-shot.

\textbf{Table-based Prompt.}
This form of prompt is designed for data that contains too many features or meaningless symbols. As their features are too many or do not have any semantic information, it is hard to describe them in natural language with limited words. Therefore, in the [prompt], we explain the data is composed of meaningless symbols and provide the number of features; in the [input] section, we directly tell the values of each data.
Therefore, it is concise and convenient to construct highly structured data. 

\textbf{Description-based Prompt.}
This form of prompt is designed for the rest datasets that have clear semantic information about the features.
Here, we use natural language in [input] to re-explain the meaning of features and the corresponding numerical values for each data. For instance, in credit scoring, we transfer the features as ``The purpose is car (new). The state of credit amount is 2366.". This form makes LLMs easier to understand the data.

Because many datasets have severe imbalance issues, we further consider the balance of the instruction tuning data. On some extremely imbalanced datasets, including Credit Card Fraud, ccFraud, Taiwan Economic Journal and Travel Insurance, we resample the minority class of the instruction tuning data. The resampling process results in a ratio of 2:1 between majority class samples and minority class samples, with a total sample size of 45k.

\vspace{-2mm}
\section{Credit and Risk Assessment Large Language Model (CALM)}
We further build our \textbf{C}redit and Risk \textbf{A}ssessment Large \textbf{L}anguage \textbf{M}odel (CALM) by fine-tuning the latest LLM, Llama2-chat \cite{touvron2023llama}, with the our instruction-tuning dataset. The instruction-tuning dataset includes all the 6 datasets that are constructed and resampled earlier. We exclude 3 datasets (the Lending Club, Polish, and PortoSeguro datasets) to verify the generalization ability.
When fine-tuning our LLM (CALM), we use the LORA strategy \cite{hu2021lora} to reduce the computation cost. The instruction tuning data is divided into a 7:1 ratio for training and validation. We set the maximum length of input texts as 2048 and fine-tune 5 epochs based on AdamW optimizer \cite{loshchilov2017decoupled} on 4 A100 40GB GPUs. We set the initial learning rate and the weight decay as 3e-4 and 1e-5, respectively. The batch size is set to 24, and the warmup steps to 1\%.

\section{Evaluation for Benchmark}

In our benchmark, we conduct a comprehensive evaluation from two aspects: model performance and bias. We show all these evaluation metrics of each dataset in our benchmark in Table \ref{tab:metric}. For model performance, we evaluated the effectiveness of the model itself as well as the significant issue of imbalance. Regarding bias, we examined the bias in the dataset and the potential bias in LLMs.

Specifically, we use the two most commonly used metrics, accuracy (Acc) and F1 score to evaluate the performance of these binary classification tasks in our benchmark. Here, we use the majority class as the positive, like most of the references.
We use the Matthews Correlation Coefficient (Mcc) as a metric \cite{chicco2020advantages} to evaluate the performance with the imbalanced situation.
Therefore, for datasets like German, Australia, and Lending Club that are relatively balanced, the F1 score reflects their overall performance. For the remaining datasets, we prioritize the Mcc metric to assess whether the model can handle class imbalance issues. When there is a certain high level of Mcc value (larger than 0), models with higher F1 and Acc scores perform better. 
We also record the Miss value of LLMs to reflect the tasks they cannot answer. A higher Miss value indicates that the LLMs' answers are more irrelevant.

To verify the bias, we follow the previous work \cite{jammalamadaka2023responsible} with the AI FAIRNESS 360 framework \cite{aif360-oct-2018}. 
We separately consider the bias of the data and the bias of the model to determine if the model has bias and if these biases are caused by the data.
For the bias of the data, we use the Disparate Impact (DI) value which computes the ratio between the probability of the unprivileged group getting a favorable prediction and the probability of the privileged group getting a favorable prediction. The DI value closer to 1 indicates a more balanced distribution, while a difference greater than 0.1 from 1 suggests potential bias risks. 
For the bias of models, we use Equal Opportunity Difference (EOD) and Average Odds Difference (AOD).
The first metric computes the difference between TPR values of unprivileged and privileged groups. 
The second metric computes the average of TPR difference and FPR difference between unprivileged and privileged groups. When the values of EOD and AOD approach 0, it indicates that the model’s
judgment is unbiased. However, when their absolute values are greater than 0.1, we should be cautious as the model may exhibit bias.
In our benchmark of bias, we take into account the potential bias that may arise from gender, age, and foreign status. We have analyzed the impact of gender, age, and foreign status on German, the impact of gender on ccFraud, and the impact of age on Travel Insurance.
We set the old, female and foreigner as the unprivileged groups for gender, age, and foreign status, respectively. We divide the age group into `young' and `old', with the age of 45 as the dividing line.

Our benchmark covers all 9 datasets from 4 tasks: credit scoring, fraud detection, financial distress identification, and claim analysis. Among these datasets, the prompts for Australia, Credit Card Fraud, PortoSeguro, Polish, and Taiwan Economic Journal are in table-based form, while the rest are in description-based form.
Except this, we introduce an additional dataset for customs fraud detection \cite{jeong2022customs}, which is similar to the aforementioned tasks, and not included in the fine-tuning process. This dataset is further evaluated (see Appendix \ref{customdata}) to test the models' generalization ability.

\begin{table}[!htb]
\vspace{-3.5mm}
  \centering
  \caption{The metrics of each dataset in our benchmark.}
\vspace{-3.5mm}
  \resizebox{0.45\textwidth}{!}{
    \begin{tabular}{cp{14.21em}}
    \toprule
    Metrics & Dataset \\
    \midrule
    Acc, F1, Miss & German, Australia, Lending Club \\
    \midrule
    Acc, Mcc, F1, Miss & Credit Card Fraud, ccFraud, Polish, Taiwan Economic Journal, PortoSeguro, Travel Insurance \\
    \midrule
    DI, EOD, AOD & German, ccFraud, Travel Insurance\\
    \bottomrule
    \end{tabular}%
}
\vspace{-3mm}
  \label{tab:metric}%
\end{table}%

\begin{table*}[!htb]
\centering
\vspace{-1mm}
\caption{The performance of LLMs and the SOTA expert system models on our benchmark. We use bold to indicate the best and underline to indicate the second-best. For Miss, where smaller is better, for other metrics, larger is better.}
\label{tab:bench}
 \vspace{-3mm}
\resizebox{0.999\textwidth}{!}{%
    \begin{threeparttable}
    \begin{tabular}{cccccccccccccc}
    \toprule
    Dataset & Data Type & Metrics & SOTA expert system models & ChatGPT & GPT4  & Bloomz & Vicuna & Llama1 & Llama2 & Llama2-chat & Chatglm2 & FinMA & CALM \\
    \midrule
    \multirow{3}[2]{*}{German} & \multirow{3}[2]{*}{Description} & Acc   & \textbf{0.804}\cite{wang2022deep} & 0.440 & 0.545 & 0.315 & 0.590 & \underline{0.660} & \underline{0.660} & 0.475 & 0.505 & 0.170 & 0.565 \\
          &       & F1    & \textbf{0.857}\cite{yao2022novel} & 0.410 & 0.513 & 0.197 & 0.505 & 0.173 & 0.173 & 0.468 & 0.477 & 0.170 & \underline{0.535} \\
          &       & Miss  & -     & 0.000 & 0.000 & 0.110 & 0.000 & 0.000 & 0.000 & 0.000 & 0.110 & 0.000 & 0.000 \\
    \midrule
    \multirow{3}[2]{*}{Australia} & \multirow{3}[2]{*}{Table} & Acc   & \textbf{0.902}\cite{wang2022deep} & 0.425 & \underline{0.748} & 0.568 & 0.489 & 0.432 & 0.432 & 0.432 & 0.115 & 0.410 & 0.518 \\
          &       & F1    & \textbf{0.875}\cite{yao2022novel} & 0.257 & \underline{0.746} & 0.412 & 0.513 & 0.412 & 0.412 & 0.260 & 0.165 & 0.410 & 0.492 \\
          &       & Miss  & -     & 0.000 & 0.000 & 0.000 & 0.000 & 0.000 & 0.000 & 0.000 & 0.806 & 0.000 & 0.000 \\
    \midrule
    \multirow{3}[2]{*}{Lending Club\tnote{*}} & \multirow{3}[2]{*}{Description} & Acc   & 0.777\cite{jemai2023feature} & 0.386 & 0.762 & 0.693 & \underline{0.808} & \underline{0.808} & \underline{0.808} & \textbf{0.809} & 0.469 & 0.572 & 0.571 \\
          &       & F1    & \textbf{0.780}\cite{jemai2023feature} & 0.401 & \underline{0.740} & 0.675 & 0.723 & 0.062 & 0.062 & 0.723 & 0.503 & 0.610 & 0.608 \\
          &       & Miss  & -     & 0.000     & 0.000 & 0.139 & 0.000     & 0.000 & 0.000    & 0.000 & 0.000  & 0.000 & 0.286 \\
    \midrule
    \multirow{4}[2]{*}{Credit Card Fraud} & \multirow{4}[2]{*}{Table} & Acc   & 0.880\cite{ileberi2022machine}\tnote{**} & \underline{0.998} & 0.810 & 0.001 & \textbf{0.999} & 0.823 & \textbf{0.999} & -     & 0.001 & 0.003 & 0.947 \\
          &       & Mcc   & \underline{0.167}\cite{asha2021credit} & -0.001 & \textbf{0.331} & 0.000 & 0.000 & -0.008 & 0.000 & -     & 0.000  & 0.002 & 0.151 \\
          &       & F1    & 0.846\cite{ileberi2022machine}\tnote{**} & \textbf{0.998} & 0.878 & 0.000 & \textbf{0.998} & 0.902 & \textbf{0.998} & -     & 0.000 & 0.004 & \underline{0.971} \\
          &       & Miss  & -     & 0.000 & 0.110 & 0.000 & 0.000 & 0.176 & 0.000 & 1.000 & 0.000 & 0.000 & 0.000 \\
    \midrule
    \multirow{4}[2]{*}{ccFraud} & \multirow{4}[2]{*}{Description} & Acc   & 0.826\cite{kamaruddin2021egrnn++} & 0.173 & 0.580 & 0.059 & 0.608 & \textbf{0.941} & \underline{0.941} & 0.914 & 0.085 & 0.060 & 0.514 \\
          &       & Mcc   & \textbf{0.344}\cite{kamaruddin2021egrnn++} & 0.066 & 0.113 & 0.000 & -0.095 & 0.000 & 0.000 & -0.020 & -0.024 & -0.061 & \underline{0.192} \\
          &       & F1    & \textbf{0.899}\cite{kamaruddin2021egrnn++} & 0.214 & 0.587 & 0.007 & \underline{0.651} & 0.007 & 0.007 & 0.437 & 0.109 & -0.006 & 0.627 \\
          &       & Miss  & -     & 0.000 & 0.210 & 0.000 & 0.000 & 0.000 & 0.000 & 0.000 & 0.891 & 0.000 & 0.000 \\
    \midrule
    \multirow{4}[2]{*}{Polish\tnote{*}} & \multirow{4}[2]{*}{Table} & Acc   & \textbf{0.968}\cite{aly2022intelligent} & 0.930 & 0.650 & 0.051 & \underline{0.949} & 0.001 & \underline{0.949} & 0.484 & 0.224 & \underline{0.949} & 0.475 \\
          &       & Mcc   & \textbf{0.569}\cite{mukeri2020financial} & 0.019 & -0.026 & 0.000 & 0.000 & 0.003 & 0.000 & \underline{0.036} & 0.010 & 0.000 & 0.015 \\
          &       & F1    & \textbf{0.986}\cite{mukeri2020financial} & 0.917 & 0.623 & 0.005 & \underline{0.924} & 0.001 & \underline{0.924} & 0.633 & 0.360 & \underline{0.924} & 0.585 \\
          &       & Miss  & -     & 0.000 & 0.000 & 0.000 & 0.000 & 0.999 & 0.000 & 0.499 & 0.761 & 0.000 & 0.000 \\
    \midrule
    \multirow{4}[2]{*}{Taiwan Economic Journal} & \multirow{4}[2]{*}{Table} & Acc   & \textbf{0.999}\cite{muslim2023ensemble} & \underline{0.968} & 0.730 & 0.032 & 0.167 & \underline{0.968} & \underline{0.968} & 0.336 & 0.396 & \underline{0.968} & 0.497 \\
          &       & Mcc   & -     & 0.000 & \textbf{0.150} & 0.000 & 0.023 & 0.000 & 0.000 & -0.016 & 0.008 & 0.000 & \underline{0.046} \\
          &       & F1    & -     &\textbf{ 0.952} & \underline{0.750} & 0.002 & 0.266 & \textbf{0.952} & \textbf{0.952} & 0.493 & 0.557 & \underline{0.968} & 0.636 \\
          &       & Miss  & -     & 0.000 & 0.010 & 0.000 & 0.644 & 0.000 & 0.000 & 0.651 & 0.593 & 0.000 \\
    \midrule
    \multirow{4}[2]{*}{PortoSeguro\tnote{*}} & \multirow{4}[2]{*}{Table} & Acc   & 0.868\cite{hanafy2021machine}\tnote{**} & \textbf{0.970} & 0.790 & 0.030 & 0.000 & -     & 0.030 & 0.049 & 0.588 & 0.050 & \underline{0.964} \\
          &       & Mcc   & \textbf{0.728}\cite{hanafy2021machine}\tnote{**} & 0.000 & -0.030 & 0.000 & \underline{0.000} & -     & 0.000 & -0.009 & -0.011 & \underline{0.008} & -0.013 \\
          &       & F1    & 0.810\cite{hanafy2021machine}\tnote{**} & \textbf{0.955} & 0.778 & 0.002 & 0.000 & -     & 0.002 & 0.040 & 0.716 & 0.040 & \underline{0.952} \\
          &       & Miss  & -     & 0.000 & 0.000 & 0.000 & 0.947 & 1.000 & 0.000 & 0.000 & 0.000 & 0.000 & 0.000 \\
    \midrule
    \multirow{4}[2]{*}{Travel Insurance} & \multirow{4}[2]{*}{Description} & Acc   & 0.839\cite{li2023exploring} & \textbf{0.981} & 0.835 & 0.015 & 0.015 & 0.000 & 0.015 & 0.665 & 0.154 & 0.002 & \underline{0.929} \\
          &       & Mcc   & \textbf{0.154}\cite{li2023exploring} & -0.008 & \underline{0.153} & 0.000 & 0.000 & -0.001 & 0.000 & 0.010 & -0.005 & 0.000 & 0.076 \\
          &       & F1    & 0.912\cite{li2023exploring} & \underline{0.975} & 0.897 & 0.000 & 0.130 & 0.001 & \textbf{0.978} & 0.787 & 0.955 &0.001 & 0.950 \\
          &       & Miss  & -     & 0.000 & 0.000 & 0.000 & 0.000 & 0.999 & 0.000 & 0.000 & 0.000 & 0.000 & 0.000 \\
    \bottomrule
    \end{tabular}
    
 \begin{tablenotes}
    \item[*] These datasets are not used to train CALM.
   \item[**] The related studies balance the data for the test set, and the values are for reference only.
    \end{tablenotes}
    \end{threeparttable}
}
 \vspace{-6mm}
\end{table*}

\section{Experiment}
In this section, extensive experiments are conducted to validate our hypotheses. Specifically, our experiments address the following questions.

\begin{itemize}
    \item \textbf{RQ1}: From our benchmark, do LLMs have the potential to generally solve the tasks of credit and risk assessment based on their pertaining process?
    \item \textbf{RQ2}: After fine-tuning LLMs with a portion of the datasets, can LLMs transfer insights between datasets and enhance their performance on untrained datasets for credit and risk assessment?
    \item \textbf{RQ3}: Is there a bias present in LLMs when it comes to solving these tasks? If so, how does this bias present?
\end{itemize}

\subsection{Experiment Setup}
\label{expset}
In addition to our model CALM, we also compare other LLMs and SOTA expert system models as follows.

We choose two kinds of the latest and most popular LLMs as the baselines, the open-resource LLMs and the non-open-resource LLMs.
For the open resource LLMs, we use (1) Bloomz \cite{muennighoff2022crosslingual}: it is capable of following human instructions in dozens of languages zero/one-shot, (2) Vicuna \cite{chiang2023vicuna}: it is an open-source chatbot trained by fine-tuning LLaMA on user-shared conversations collected from ShareGPT, (3) Llama1 \cite{touvron2023llama}: it only uses publicly available data and has competitive performance compared to the best existing LLMs, (4) Llama2 \cite{touvron2023llama}: the new version of Llama, (5) Llama2-chat \cite{touvron2023llama}: it is further optimized for dialogue use cases, (6) Chatglm2 \cite{du2022glm}: it is the second-generation version of the open-source bilingual (Chinese-English) chat model ChatGLM-6B that has stronger math ability. (7) FinMA (7B-full) \cite{xie2023pixiu}: It is the latest LLMs fine-tuned in the financial field from the PIXIU project. To ensure fairness and minimize computation cost, we use the around 7B-parameters version for all these LLMs.
For the non-open resource LLMs, we use (1) ChatGPT: A powerful LLM from OpenAI; (2) GPT-4 \cite{openai2023gpt4}: A powerful LLM with around 1T parameters proposed by OpenAI. 

In addition, we have also included a comparison of the results from the SOTA expert system models on these datasets in Table \ref{tab:bench}. Most of the models are tree-based or deep neural networks, which are trained for specific tasks.

\vspace{-2mm}
\subsection{Results}
\subsubsection{\textbf{The benchmark of Credit and Risk Assessment (RQ1).}} Table \ref{tab:bench} shows the results of the benchmark with the LLMs and SOTA expert systems.
Overall, as for performance, we can find that GPT-4 may have the ability to solve these tasks at the same time, even close to some SOTA expert systems in some tasks (like Lending Club and Travel Insurance), but the other LLMs still have a certain gap. The results also indicate that LLMs can be divided into four groups, non-open-source LLMs (ChatGPT and GPT 4), open-source LLMs (Bloomz, Vicuna, Llama1, and Llama2), open-source LLMs with a chat version (Llama2-chat and Chatglm 2), and financial LLMs (FinMA), each with its distinct characteristics.

For the non-open-source LLMs, we observe that ChatGPT and GPT-4 possess capabilities that are comparable to SOTA expert systems, especially GPT-4. GPT-4 demonstrates exceptional handling of imbalanced data, achieving optimal or near-optimal results on several datasets. Additionally, GPT-4 may also have the ability to handle meaningless data, such as the Australia.
These impressive results are noteworthy considering that our prompts are only provided as one-shot and even nearly zero-shot.
Furthermore, these findings address our 
\textbf{RQ1}. 
ChatGPT and GPT-4 can acquire strong generalization abilities to tackle multiple tasks specific to credit and risk assessment in the financial domain without further supervised training. This sets them apart from these SOTA expert systems listed in Table \ref{tab:bench}, which exhibit excellent performance on individual datasets, but lack versatility and cannot be applied to other tasks.

However, for the open-source LLMs (Bloomz, Vicuna, Llama1, and Llama2), the Acc, F1 (both of them near the class ratio) and Mcc (most are around zero) show that most of them tend to predict all the samples to one class, regardless of whether the dataset is balanced or imbalanced. 
This indicates that they can only answer the question but lack reasoning ability. Among these LLMs, Bloomz and Llama1,2 give opposite predictions in some datasets, such as Taiwan Economic Journal, Polish and ccFraud, which may be due to their training data, making them learn the different answers. 

Interestingly, for the open-source LLMs with a chat version, most Mcc of Llama2-chat and Chatglm2 are not equal to zero. This may be because they are further trained on the conversation data, which makes them try to give a reasonable answer. 
However, this also makes them cannot give direct answers to our questions (or refuse to answer), showing some higher Miss values.
In addition, their ability is still insufficient, which gives predictions in the opposite direction in some tasks (Mcc$<$0).

The FinMA (Financial LLM) has demonstrated its strong ability to address issues, due to its extensive training in financial tasks. In addition, in comparison to open-source LLMs with a chat version, although the accuracy and F1 scores of FinMA are not higher than theirs, the Mcc value (Mcc does not equal 0 in Credit Card Fraud and PortoSeguro) indicates that FinMA may be capable of providing discerning results based on previously trained financial tasks.

Therefore, all the results indicate that LLMs have developed a general ability to understand different credit and risk assessment tasks, but the performance of these answers may vary depending on the training data and parameter size of the LLMs.
GPT-4 is approaching the power of expert systems, which may provide a new chance for solving these tasks.
However, these open-source general or financial LLMs that are not specifically trained on credit and risk assessment may still lack this solving ability. 

\vspace{-2mm}
\subsubsection{\textbf{The fine-tuned results of our LLM (RQ2).}}
\begin{figure}[!htb]
\vspace{-4mm}
\centering
\includegraphics[width=0.95\columnwidth]{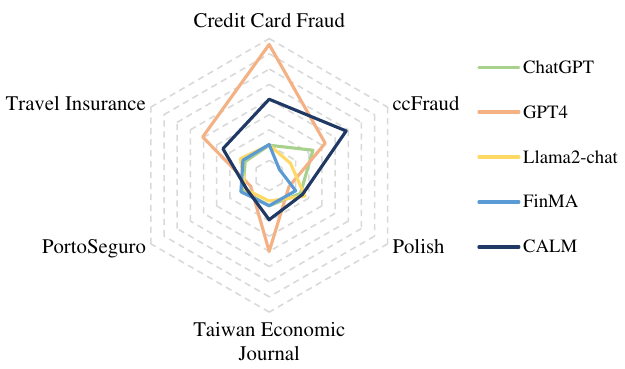}
\vspace{-3mm}
\caption{The radar figure for Mcc metric on different LLMs. The outermost value is 0.35 and it decreases by 0.05 for each subsequent layer.}
\vspace{-4.5mm}
\label{fig-radar}
\end{figure}

The results of our LLM (CALM) are shown in the last column of Table \ref{tab:bench}. Because of the significant proportion of imbalanced data in our benchmark, we also create a radar figure for Mcc metric to visually display the abilities of CALM, original Llama2-chat, ChatGPT, and GPT-4, shown as Figure \ref{fig-radar}.

After fine-tuning, our CALM demonstrates capabilities that are comparable to those of GPT-4. It exhibits the ability to learn credit tasks through instruction tuning, even on some untrained datasets.
On the datasets we trained like Credit Card Fraud, ccFraud and Taiwan Economic Journal, CALM has a higher value of Mcc than before and even better than ChatGPT and GPT-4. 
It means that CALM can make predictions based on what it has learned from the training set, rather than guessing.
In a more general sense, the Mcc value of CALM typically increases, though there may be a slight decrease in Acc and F1 for some datasets.  
This comes from that CALM changes the outcomes from some majority-class samples to more balanced results.

When testing CALM on untrained data from lending club, Polish, and PortoSeguro, we find potential for generalization in similar datasets but unclear on some other datasets.
CALM shows improvement in understanding and answering questions on the Polish dataset, with a Miss value of 0. However, it does not achieve significant results on the Lending Club and PortoSeguro datasets. This could be due to the similarity of Polish to other training data, while PortoSeguro consists of anonymized data and requires specialized learning.
In FinMA, a similar phenomenon is observed, where the training on a broader set of financial tasks unrelated to the task in Figure \ref{fig-radar} results in better performance in PortoSeguro compared to its base model (llama2-chat). This also suggests potential commonalities among financial tasks.
To delve deeper, we further use an additional task and corresponding dataset to test whether fine-tuning LLMs on certain tasks can improve their ability on other tasks, shown in Appendix \ref{customdata}. The results provide a clearer result of the transferrable skills possessed by LLMs.

Therefore, the results indicate that LLMs with fine-tuning have great potential in solving different credit and risk assessment tasks. 
It demonstrates that LLMs may be capable of learning from the related data and have the extrapolation ability without requiring special construction for each dataset like expert systems, which echoes our \textbf{H2}.
It will be feasible for companies or individuals to create a customized LLM that aligns with their needs quickly.

\subsubsection{\textbf{The bias analyze (RQ3).}} 
We explore the bias of three LLMs (ChatGPT, GPT-4, and our model CALM) in three datasets. Our results indicate that the inherent biases present in these datasets are relatively small. However, it is crucial to acknowledge that there is a notable risk of bias for LLMs.

\begin{figure}[!htb]
\vspace{-3mm}
\centering
\includegraphics[width=0.95\columnwidth]{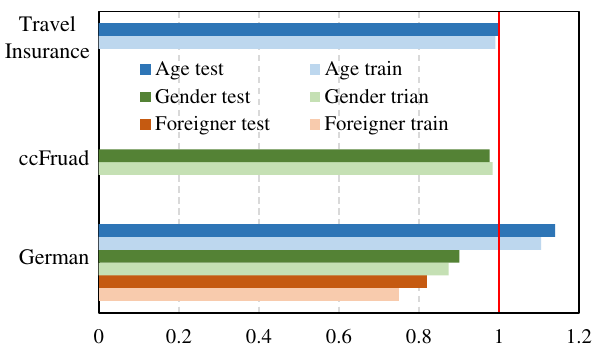}
\vspace{-4mm}
\caption{The Disparate Impact (DI) value of the train/test data on three datasets. Closer to 1 is better.}
\vspace{-5mm}
\label{fig-bias}
\end{figure}

For potential biases within each dataset, we analyze whether the benchmark and instruction-tuning data themselves have any bias. 
We consider the impact of gender, age, and foreign status on German, the impact of gender on ccFraud, and the impact of age on Travel Insurance. The fundamental bias information of these datasets can be seen in Figure \ref{fig-bias}.
We can find that except for the `foreigner' in German, all of these DI values are near 1.
This suggests that the majority of the original datasets are unbiased when it comes to these sensitive features. Additionally, instruction tuning itself is unbiased towards the model.

To evaluate the bias of LLMs, we calculate the Equal Opportunity Difference (EOD) and the Average Odds Difference (AOD) on these features with the predictions made by LLMs.
The results are shown in Table \ref{tab:bias}. For ChatGPT and GPT-4, it indicates that they have a bias in some special cases. For example, GPT-4 is more likely to give females wrong predictions (AOD is -0.273) on the ccFraud dataset and prefer foreign workers on the German dataset (EOD is 0.289), even though the original test data is unbiased (DI close to 1); on the German dataset, ChatGPT prefers to lend money to older people (EOD is 0.137). It's also interesting to note that the potential biases that exist in both ChatGPT and GPT-4 are not completely consistent with each other (`gender' and `age' in German, and `gender' in ccFraud). 
This may be related to their training dataset and the alignment process of reinforcement learning human feedback \cite{casper2023open,wolf2023fundamental}.
For our CALM, the risk of bias is similar to ChatGPT and GPT-4, with the greatest bias found in the foreigner of German.
This is because the other data is more balanced, so our LLM trained with these data does not suffer such bias. However, the foreigner variable in German has a tendency (DI$=0.75$ in training data), which causes CALM to learn the bias.

In summary, we demonstrate that there is a potential bias risk in credit and risk assessment with LLMs (GPT-4 is more serious than ChatGPT). Referring back to our \textbf{H3}, if individuals and companies promote the application of LLM in these tasks within society, the issue of bias needs to be addressed.

\begin{table}[!htb]
\vspace{-2mm}
  \centering
  \caption{The bias evaluation of ChatGPT, GPT4, and our LLM on the test data. We use bold to indicate the best and underline to indicate the second-best. Closer to 0 is better.}
  \vspace{-3mm}
  \resizebox{0.48\textwidth}{!}{
  \begin{tabular}{ccccccc}
    \toprule
    \multirow{2}[4]{*}{Metrics} & \multirow{2}[4]{*}{Model} & \multicolumn{3}{c}{German} & ccFraud & Travel Insurance \\
\cmidrule{3-7}          &       & Gender & Age   & Foreigner & Gender & Age \\
    \midrule
    \multirow{3}[2]{*}{EOD} & ChatGPT & \underline{0.101} & 0.137 & \textbf{-0.023} & \textbf{0.001} & \textbf{0.001} \\
          & GPT4  & \textbf{0.010} & \underline{-0.104} & 0.289 & -0.023 & \underline{0.007} \\
          & CALM & -0.121 & \textbf{-0.075} & \underline{-0.250} & \underline{0.003} & \underline{0.007} \\
    \midrule
    \multirow{3}[2]{*}{AOD} & ChatGPT & \textbf{0.005} & 0.108 & \underline{0.093} & \textbf{0.036} & \textbf{0.000} \\
          & GPT4  & -0.156 & \textbf{-0.092} & \textbf{0.040} & -0.273 & -0.115 \\
          & CALM & \underline{-0.116} & \underline{-0.103} & -0.207 & \underline{0.065} & \underline{0.055} \\
    \bottomrule
    \end{tabular}
}
\vspace{-5mm}
  \label{tab:bias}%
\end{table}%


\section{Conclusion}
In this work, we explore the impact of LLMs on online credit and risk assessment and highlight the potential implications of the bias of LLMs\footnote{We provide our ethics statement and limitations in Appendix \ref{eth} and \ref{lim}, respectively.}.
To test the effectiveness of LLMs in credit and risk assessment, we build the first benchmark, open source LLM (CALM), and evaluate LLMs' performance against existing expert systems on our build benchmark. 
We observe that although existing open-source Language Models (LLMs) may not be able to process financial tabular data without modifications, their pretraining endows them with a powerful understanding capability.
Furthermore, GPT-4 or LLMs fine-tuned with more relevant data have the potential to achieve superior results and potentially replace existing expert systems.
However, through our bias experiences, we also discover that LLMs such as ChatGPT or GPT-4 exhibit biases, affecting individuals' access to online financial services and opportunities. As LLMs continue to evolve and become more prevalent, it is essential to address their limitations and biases to ensure fair and unbiased decision-making.


\newpage
\bibliographystyle{ACM-Reference-Format}
\bibliography{ref}

\newpage
\appendix

\section{More Results}
\label{customdata}
Here, we evaluate the potential of LLMs in customs fraud detection dataset \cite{jeong2022customs} which is similar to but not strictly belonging to credit and risk assessment. This is aimed at showing that LLMs can not only complete credit and risk assessments but also be applied in a broader field of risk detection for the whole society, which demonstrates LLMs can potentially recognize and apply transferrable skills.

\textit{CustomsDeclaration\footnote{\url{https://github.com/Seondong/Customs-Declaration-Datasets}}}
 comprises customs import declaration records and is intended to detect fraudulent attempts to reduce customs duty or critical frauds that can threaten public safety. 
It consists of 54,000 artificially generated records created by CTGAN with 24.7 million customs declarations reported from January 1, 2020, to June 30, 2021.
The dataset encompasses 20 attributes and includes two labels: ``fraud" and ``critical fraud". 
The label ``fraud" involves binary classification, which aims to detect fraudulent attempts to reduce customs duty with ``non-fraud" and ``fraud" two cases. 
The label ``critical fraud" involves three classes: ``non-fraud", ``fraud", and ``critical fraud", making it a multi-class classification task. It aims to detect critical frauds that can threaten public safety.
Due to the extremely imbalanced issue of the label ``critical fraud", which is also not used in the original paper, we use ``fraud" as the prediction target.
To evaluate the dataset, we sample 2000 instances from their test set to make the prompt data.

\textit{Customs}:
\begin{quote}\begin{scriptsize}\begin{verbatim}
{"id": 0,
"Prompt": "Identify the provided customs import 
declaration information to determine whether it 
constitutes customs fraud that attempts to 
reduce customs duty or not. The answer 
must be 'no' or 'yes', and do not provide any 
additional information. This Import Declaration 
consists of 20 data attributes, including 
Declaration ID, Date, Office ID, Process type, 
Import type, Import use, Payment type, Mode of 
transport, Declarant ID, Importer ID, Seller ID, 
Courier ID, HS6 code, Country of departure, 
Country of origin, Tax rate, Tax type, Country of 
origin indicator, Net mass and Item price. For 
instance, 'This customs import declaration has 
attributes: Declaration ID: 97061800, Date: 
2020-01-01, Office ID: 30, Process Type: B, ...,
Item Price: 372254.4.' should be categorized as 
'no'.",
"Text": "This customs import declaration has 
attributes: Declaration ID: 97061800, Date: 
2020-01-01, Office ID: 30, Process Type: B, 
Import Type: 11, Import Use: 21, Payment Type: 
11, Mode of Transport: 10, Declarant ID: ZZR1LT6, 
Importer ID: QLRUBN9, Seller ID: 0VKY2BR, Courier 
ID: nan, HS6 Code: 440890, Country of Departure: 
BE, Country of Origin: BE, Tax Rate: 0.0, Tax 
Type: FEU1, Country of Origin Indicator: G, 
Net Mass: 108.0, Item Price: 372254.4. Answer:",
"Answer": "no", 
"Choices": ["no", "yes"], 
"Gold": 0}
\end{verbatim}\end{scriptsize}\end{quote}


\textbf{Metrics.} Except for the Acc and F1, we also use the precision (P). Due to the reason that the LLMs cannot give a probability, we are not able to compute precision@n\% as the original paper. We also consider `fraud' as a positive class, like the original paper.

\begin{table}[!htb]
  \centering
  \caption{The results of Customs dataset.}
  \resizebox{0.45\textwidth}{!}{
  \begin{threeparttable}
        \begin{tabular}{ccccc}
    \toprule
          & Acc   & P     & F1    & Miss \\
    \midrule
    SOTA expert system & -     & \textbf{0.646}\cite{jeong2022customs}\tnote{*} & -     & - \\
    ChatGPT & \textbf{0.778} & 0.000 & 0.000 & - \\
    GPT4  & \underline{0.665} & \underline{0.255} & 0.264 & - \\
    Bloomz & 0.222 & 0.222 & \underline{0.363} & - \\
    Vicuna & \textbf{0.778} & 0.000 & 0.000 & - \\
    Llama1 & \textbf{0.778} & 0.000 & 0.000 & 0.001 \\
    Llama2 & \textbf{0.778} & 0.000 & 0.000 &  - \\
    Llama2-chat & 0.290 & 0.214 & 0.339 & - \\
    Chatglm2 & 0.005 & 0.000 & 0.000 & 0.994 \\
    CALM & 0.303 & 0.230 & \textbf{0.368} & 0.001 \\
    \bottomrule
    \end{tabular}%
    \begin{tablenotes}
    \item[*] It is the value of precision@10\%.
    \end{tablenotes}
    \end{threeparttable}
}
  \label{tab:customs}%
\end{table}%

\textbf{Results.} We show the results in Table \ref{tab:customs}.
Based on these results, we can reach a conclusion similar to our benchmark. GPT4 has great potential in solving related tasks. However, the open-source LLMs and even ChatGPT may lack this ability. In particular, due to the reversal of the definition of positive class, we can clearly see that the original open-source LLMs and  ChatGPT are completely biased towards one side for prediction. For example, although Acc is 0.778, the precision and F1 of Vicuna, Llama1,2 and ChatGPT equal 0. Nevertheless, through training on similar tasks, LLMs may learn the ability to generalize predictions. In particular, our 7B-CALM that only be trained on credit and risk assessment has higher precision and F1 compared to other open-source LLMs and can even outperform the ChatGPT.

More importantly, the results not only suggest that LLMs have the potential for further application in credit and risk assessment, but also in a broader range of social activities related to credit and risk. This indicates that the transferrable skills of LLMs.

\section{Ethics Statement}
\label{eth}
This research investigates the effectiveness of LLMs in credit and risk assessment with some online tasks. 
The datasets used for this research were constructed based on online open-source data that can be further modified. We have thoroughly reviewed the data to ensure that it does not contain any personally identifiable or offensive information. Therefore, we are confident that the datasets are safe and suitable for distribution.

\section{Limitations}
\label{lim}
Although this research has several contributions, we acknowledge two limitations. First, due to computational constraints, we test and fine-tune open-resource LLMs with a smaller parameter size, which may underestimate the potential of the LLMs. Second, although the LLMs have strong language explanatory abilities, we do not construct an interpretable dataset, which may increase the interpretability. The reason is that it requires a significant amount of additional professional labeling.

\end{document}